\documentclass[letterpaper, 10 pt, conference]{ieeeconf}  

\IEEEoverridecommandlockouts                              
\usepackage{graphicx}
\usepackage{wrapfig} 
\usepackage{color}
\usepackage{amsmath}
\usepackage{amssymb}
\usepackage{bm}
\usepackage{dsfont}
\usepackage{tabularx}
\usepackage{tabularray}
\usepackage{hyperref} 
\hypersetup{  
    colorlinks=true,  
    linkcolor=blue,  
    filecolor=magenta,        
    urlcolor=cyan,  
    citecolor=blue  
} 

\title{\LARGE \bf
ZSL-RPPO: Zero-Shot Learning for Quadrupedal Locomotion in Challenging Terrains using Recurrent Proximal Policy Optimization
}

\author{Yao Zhao$^{1\dagger}$, Tao Wu$^{2\dagger}$, Yijie Zhu$^{1}$, Xiang Lu$^{1}$,\\
Jun Wang$^{3}$, Haitham Bou-Ammar$^{4}$, Xinyu Zhang$^{5}$, Peng Du$^{1\ast}$\\
\thanks{$^{1}$Yao Zhao, Yijie Zhu, Xiang Lu and Peng Du are with Huawei Technologies, China.}%
\thanks{$^{2}$Tao Wu is with Huawei Munich Research Center, Germany.}%
\thanks{$^{3}$Jun Wang is with University College London, U.K.}%
\thanks{$^{4}$Haitham Bou-Ammar is with Huawei Noah's Ark Lab, U.K.}%
\thanks{$^{5}$Xinyu Zhang is with East China Normal University, Shanghai, China.}%
\thanks{{$^\dagger$Equal contributions}.} 
\thanks{{$^\ast$Corresponding author: dp@zju.edu.cn}}
}

\newcommand{\tabincell}[2]{\begin{tabular}{@{}#1@{} }#2 \end{tabular}}
\begin{document}

\maketitle
\thispagestyle{empty}
\pagestyle{empty}

\begin{abstract}

We present ZSL-RPPO, an improved zero-shot learning architecture that overcomes the limitations of teacher-student neural networks and enables generating robust, reliable, and versatile locomotion for quadrupedal robots in challenging terrains. We propose a new algorithm RPPO (Recurrent Proximal Policy Optimization) that directly trains recurrent neural network in partially observable environments and results in more robust training using domain randomization. Our locomotion controller supports extensive perturbation across simulation-to-reality transfer for both intrinsic and extrinsic physical parameters without further fine-tuning. This can avoid the significant decline of student’s performance during simulation-to-reality transfer and therefore enhance the robustness and generalization of the locomotion controller. We deployed our controller on the Unitree A1 and Aliengo robots in real environment and exteroceptive perception is provided by either a solid-state Lidar or a depth camera. Our locomotion controller was tested in various challenging terrains like slippery surfaces, Grassy Terrain, and stairs. Our experiment results and comparison show that our approach significantly outperforms the state-of-the-art.

\end{abstract}

\section{INTRODUCTION}

Quadruped robots are more attractive compared to wheeled robotic systems, promising practical applications in rugged terrains, challenging and disorderly environments. In recent years, quadruped robot is widely studied in search and rescue operations, industrial inspection and academic research.

The complexity of positioning and planning algorithms for quadruped robots is not significantly different from that of wheeled robots. The main challenge lies in the locomotion control algorithms. Conventional locomotion control for quadruped robots adopts the Model Predictive Control (MPC) algorithm. Over the past two years, reinforcement learning algorithms have been widely applied to the locomotion control of quadruped robots, achieving impressive results. The mainstream approaches based on RL in the literature follow a teacher-student imitation training paradigm \cite{kumar2021rma,MLHWKH22}. That is, a teacher policy is first trained with access to privileged information and, in the second step, a student policy learns to imitate the teacher. However, imitation learning ends up with a wide gap between teacher and student policies and catastrophic degradation in student's real-world performance.


Our method for zero-shot simulation-to-reality transfer does not require a teacher-student framework. We present a new training paradigm of recurrent proximal policy optimization (RPPO) that learns appropriately in partially observable environments. Furthermore, we present a zero-shot transferable learning architecture that supports extensive perturbations of intrinsic and extrinsic physical parameters across sim-to-real transfer. We attached a compact depth camera/solid-state Lidar to the robot during deployment for terrain perception and reconstruction. Our results indicate that the proposed technique can transfer zero-shot to the real world and outperform state-of-the-art methods, especially in challenging terrains.

Aiming at an end-to-end deep control policy, we notice that domain randomization/adaptation and generalization issues in deep reinforcement learning are inherently partially observable. First, the details of task parameters (e.g., simulation parameters) are hidden from our learner to enable robust policies across such invariances. Precisely, the lack of task specification parameters in the inputs to the policy is helpful to allow agents to generalize beyond mistakes in hardware calibration and terrain information. Second, our method learns visuomotor action selection rules end-to-end from albeit limited sensory feedback. Consequently, policies produce actions after conditioning on Lidar or camera observations without knowing the actual low-dimensional state of the system. To this end, we present a recurrent proximal policy optimization (RPPO) approach that can directly train recurrent network architectures under partially observable environments and therefore facilitate more stable and scalable training under domain randomization than teacher-student baselines. 

Furthermore, we propose scalable domain randomization to enable zero-shot simulation-to-reality transfer in our setting. In the simulation, we jointly randomize intrinsic and extrinsic physical parameters (mass offset, the center of mass and onboard sensor parameters, if any contact friction and restitution coefficients), actuator parameters (motor strengths and PD coefficients), proprioceptive observations (noises on base linear and angular velocities, degree-of-freedom positions) and exteroceptive observations (e.g., noise on depth map or elevation map samples). In this way, our randomization covers broader task distributions enabling more robust and generalizable policies.

\section{RELATED WORK}
\subsection{A brief survey of quadruped robot control} 
The control of legged robotics is well-studied with many proposed solutions and architectures \cite{hwangbo2019learning, bjelonic2023learning, calandra2016bayesian}. One of the most prevalent variants is a modular controller design that breaks down the overall control problem into smaller, manageable, and largely decoupled systems that form a hierarchy. Each system spawns reference values to the lower component in the hierarchy that terminates with a proportional-integral-derivative (PID) controller to track generated foot stance trajectories \cite{full1999templates, raibert1981hybrid}. While leading to impressive results in rough terrain locomotion \cite{bellicoso2018advances, frey2023fast}, such modular approaches suffer from substantial drawbacks, especially regarding this model's accuracy and the laborious demand on expert designers.    


Reinforcement learning (RL) techniques have the potential to overcome those limitations by enabling agents to learn directly from data through trial and error \cite{sutton2018reinforcement}. In RL, agents gather experience using an action-selection rule (or policy/controller) that selects an action when conditioned on a state from the environment. Given this experience, the objective of an RL agent is to automatically tune the policy to maximize a reward signal that prescribes the task. In its most general form, RL algorithms are model-free stipulating little to nothing regarding the transition dynamics of the environment. As such, the process can be fully automated, optimizing controllers end-to-end from sensory data. While promising, model-free RL algorithms require many agent-environment interactions - weeks or months \cite{yu2018towards,yarats2021improving} - before discovering acceptable policies. Hence learning directly on real hardware with RL is a formidable challenge gaining much attention lately \cite{smith2022walk, cowen2022samba, rudin2022learning}. However, such approaches demand significant prior knowledge or computing power and have only been applied to relatively simple and stable platforms or when we can access accurate simulators \cite{polydoros2017survey,xu2019learning,ugurlu2021reinforcement}.

\subsection{Simulation-to-reality transfer} Rather than learning directly on robotic platforms, applying RL to legged robotics primarily involves designing simulation policies and transferring those discovered policies to the real world. To tackle the simulation-to-reality gaps, we can generally differentiate two main streams of work that aim to improve simulators to enhance simulations by performing system identification on data collected from the real robot\cite{hwangbo2019learning}, while the second accepts simulation flaws and tries to discover robust policies that are more likely to transfer to reality \cite{ju2022transferring,hofer2020perspectives,hofer2021sim2real,muratore2022neural}. As detailed in \cite{MLHWKH22}, one can achieve robust policies by randomizing essential aspects of the simulator, e.g., by perturbing the system dynamics or inducing noise in state or action representations \cite{tobin2017domain,mehta2020active,muratore2022neural,li2021reinforcement}. 

Although many robust techniques exist \cite{tessler2019action,moos2022robust,smirnova2019distributionally,abdullah2019wasserstein,hou2020robust}, deploying simulation-trained policies on quadruped robotics remains an open problem in challenging terrains. While simulation-to-reality transfer assumes a reality gap, almost all those strategies require well-behaved simulators to deploy policies successfully in the real world. 

\begin{figure*}[htbp]
  \centering  
  \includegraphics[width=\textwidth]{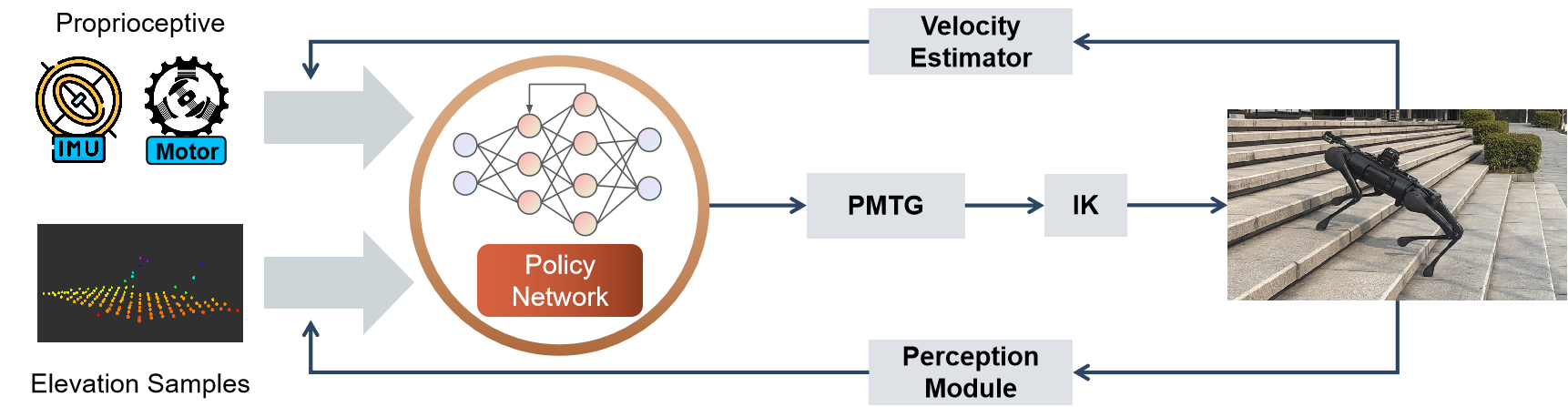} 
  \caption{Overview of the locomotion control pipeline.}  
  \label{fig:overview}  
\end{figure*}

In other attempts to demonstrate zero-shot generalization from simulation to reality, works \cite{LHWKH20}, \cite{kumar2021rma}, \cite{agarwal2023legged} can be seen as introducing a teacher-student reinforcement learning setup. Here, the teacher first learns in simulation while having access to privileged information (e.g., ground-truth knowledge of the terrain and robot's contact in \cite{LHWKH20}). The student, i.e., the real robot, then learns to imitate the teacher but only using its available sensors without accessing additional information. This direction led to impressive real-world behavior in complex outdoor terrains, effectively executing locomotion in moss, mud and vegetation environments. Motivated by these results, we also attempted teacher-student designs in our setting. However, we noticed that the student's imitation learning step that is used to learn from the teacher experienced many failures, significantly degrading the student's policy compared to the teacher's in case the simulation-to-reality gap is extensive.
\section{METHOD}
\subsection{System Overview}
The overall control pipeline is laid out in Fig. \ref{fig:overview}. The policy accepts proprioceptive measurements and exteroceptive observations to emit a distribution over the gait schedule parameters via a control policy network, paired with a parametric policy modulated trajectory generator (PMTG) and inverse kinematics (IK) for quadrupedal gaits. Gait parameters are further processed to generate valid control signals for the actuators of the robot. The neural network is trained in simulation using Deep RL, after which it is transferred to the real robot without any further training or adaptation. During deployment, velocities are estimated using a Kalman filter from the data of motor encoders and IMU. Furthermore, we provide terrain elevation samples based on a single solid-state Lidar or a single depth camera as external perception.

\subsection{Recurrent Proximal Policy Optimization}
To overcome the performance gap from the teacher-student learning paradigm, we propose a more direct approach, termed \emph{recurrent proximal policy optimization} (RPPO). 

Reinforcement learning (RL) of a control policy is abstracted as a Partially Observed Markov Decision Process (POMDP) \cite{Ast65}, with
latent states $(s_t)\in\mathcal{S}$, emitted observations $(o_t)\in\mathcal{O}$, actions $(a_t)\in\mathcal{A}$, 
state transition kernel $T:\mathcal{S}\times\mathcal{A}\to\mathcal{P}(\mathcal{S})$, emission function $E:\mathcal{S}\to\mathcal{P}(\mathcal{O})$, reward function $R:\mathcal{S}\times\mathcal{A}\to\mathbb{R}$, and reward discount $\gamma\in[0,1)$. The agent in the POMDP follows a recurrent policy $\pi_\theta(\cdot|\cdot,h):\mathcal{O}\to\mathcal{P}(\mathcal{A})$ 
with an internal state $h\in\mathcal{H}$, and interacts with the environment in an episodic setting. The episodic trajectory $\tau$ following policy $\pi_\theta$ is collected as 
\begin{equation*}
    \tau=\Big( (s_0,o_0,h_0,a_0,r_0), ..., (s_{|\tau|},o_{|\tau|},h_{|\tau|},a_{|\tau|},r_{|\tau|})\Big).
\end{equation*}
The RL objective is to maximize the expected cumulative reward $r$:
\begin{equation*}
    \text{maximize} ~ \mathbb{E}_{\tau\sim\pi_\theta}\bigg[ \sum_{t=0}^{|\tau|}\gamma^{t}r_t \bigg].
\end{equation*}

RPPO algorithm trains recurrent policies under episodic POMDP. Likewise in proximal policy optimization (PPO) \cite{SWDRK17}, policy $\pi_\theta$ is updated via optimizing an $\varepsilon$-clipped surrogate loss:
\begin{align*}
    L_\text{sur}(\theta;\tau) &= -\frac{1}{|\tau|}\sum_{t=0}^{|\tau|-1} \min \bigg( \frac{\pi_\theta(a_t|o_t,h_t)}{\pi_{\theta_\text{old}}(a_t|o_t,h_t)}\widehat A_t, \\
    &\mathrm{clip}\Big(\frac{\pi_\theta(a_t|o_t,h_t)}{\pi_{\theta_\text{old}}(a_t|o_t,h_t)};1-\varepsilon,1+\varepsilon\Big)\widehat A_t \bigg).
\end{align*}
Here $\widehat A_t$ is the advantage evaluated by generalized advantage estimation (GAE) \cite{SMLJA16}:
\begin{align*}
    \widehat Q_t^{(k)} &= \sum_{l=t}^{t+k-1}\gamma^{l-t}r_l + \gamma^k V^\pi(s_{t+k}), \\
    &\quad k=1,2,...,|\tau|-t-1, \\
    \widehat Q_t^{(|\tau|-t)} &= \sum_{l=t}^{|\tau|}\gamma^{l-t}r_l =: \widehat R_t, \\
    \widehat A_t &= -V^\pi(s_t) + \frac{1-\lambda}{1-\lambda^{|\tau|-t}}\sum_{k=1}^{|\tau|-t}\lambda^{k-1} \widehat Q_t^{(k)},
\end{align*}
where $\lambda\in(0,1)$ is a discount factor and $V^\pi:\mathcal{S}\to \mathbb{R}$ is the reward-to-go following policy $\pi$.

Implementation-wise, we approximate $V^\pi$ with a parameterized value (critic) network $V_\eta$, and update $V^\pi$ via minimizing an $\varepsilon$-clipped value loss:
\begin{align*}
  L_\text{val}(\eta;\tau) &= \frac{1}{|\tau|}\sum_{t=0}^{|\tau|-1} \max(|V_{\eta_\text{old}}(s_t) + 
  \mathrm{clip}(V_{\eta}(s_t) \\
  &-V_{\eta_\text{old}}(s_t); 
  -\varepsilon,\varepsilon) - \widehat R_t|^2, |V_{\eta}(s_t)-\widehat R_t|^2).
\end{align*}

Since $V_\eta$ is only used during the training stage (but not deployment), it is viable to leverage asymmetry between critic $V_\eta$ and actor $\pi_\theta$, in the sense that $V_\eta$ is allowed to access latent states $(s_t)$ from the simulator while $\pi_\theta$ only has access to noisy observations $(o_t)$.

\subsection{Gated Recurrent Policy Network}
\begin{figure}
  \centering  
  \includegraphics[width=0.48\textwidth]{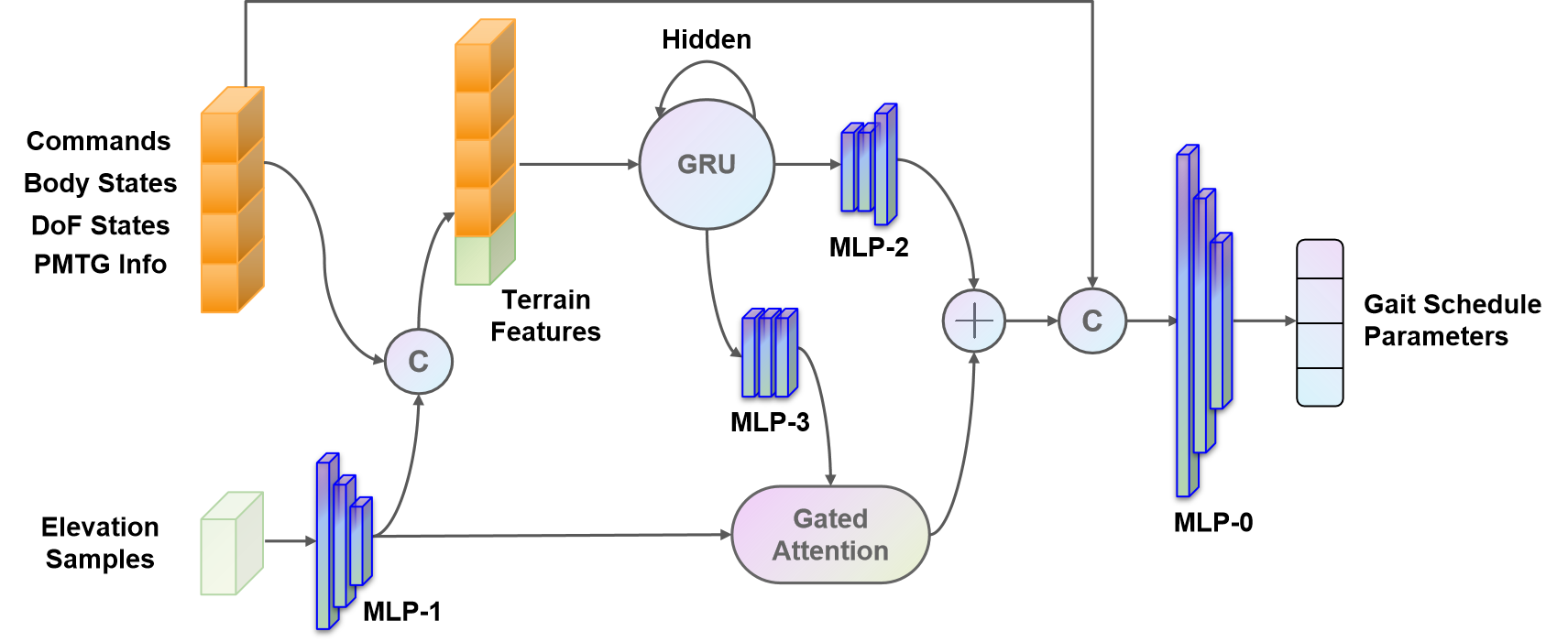} 
  \caption{Policy network architecture.}  
  \label{fig:policy_network}  
\end{figure}

We present a gated recurrent policy network (GRPN) as shown in Fig. \ref{fig:policy_network}, which has demonstrated self-adaptive solid capability in incorporating exteroception. At the core of GRPN, Gated Recurrent Unit (GRU) and gated attention \cite{MLHWKH22} are incorporated into a recurrent belief encoder.
The input size, hidden size, layer size of GRU is 197, 64 and 2. The layer size of MLP-0, MLP-1, MLP-2 and MLP-3 is (221, 256, 192, 128, 16), (187, 128, 96, 64), (64, 64, 64, 88) and (64, 64, 64, 64). Compared to alternative architectures such as temporal convolution networks \cite{BKK18}, GRPN can incorporate longer observational histories with less memory footprint. 

\begin{figure*}[htbp]
  \centering  
  \includegraphics[width=\textwidth]{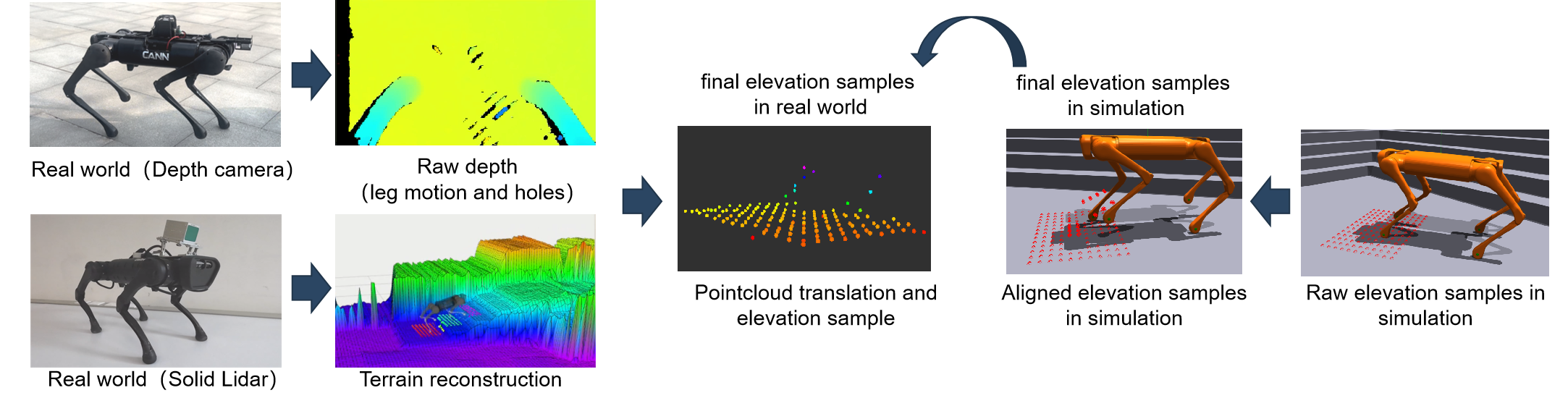} 
  \caption{We narrow the perception gap between simulation and the real world by applying pre-processing techniques and 3D reconstruction.}  
  \label{fig:perception}  
\end{figure*}  

\subsection{Observation and Action Space}
\paragraph*{Proprioception}
The input of the policy network consists of proprioceptive and exteroceptive observations.
The proprioceptive observations as shown in Table \ref{table:obs_1} include user-specified velocity command, body pose and velocities, dof positions and velocities (plus histories), dof target positions (plus histories), and PMTG phase information. These observations are directly accessible in simulation as well as reliably measurable via onboard sensors in reality.
\begin{table}[htbp]  
  \centering  
  \caption{Proprioceptive observations.}  
\begin{tabular}{cc}
    \hline
    Observation & Dim. \\
    \hline
    velocity command & 3 \\
    body pose & 3 \\ 
    base linear velocity & 3 \\
    base angular velocity & 3 \\
    dof position & 12 \\
    dof velocity & 12 \\
    dof position history (3 timesteps) & 36 \\
    dof velocity history (2 timesteps) & 24 \\
    dof target history (2 timesteps) & 24 \\
    PMTG phase information & 13 \\
    \hline
    total dimensions & 133 \\
    \hline
\end{tabular} \\[4pt]
  \label{table:obs_1} 
\end{table}

\paragraph*{Exteroception}
The exteroceptive observations are formatted as a rectangle-shaped point grid covering the base and footholds of the legged robot with extra margins. As shown in Fig. \ref{fig:perception}, for the depth camera, we sample elevation data from the area beneath the head and front legs of the robot. For the terrain mapping based on Lidar odometry, we sample the area directly beneath the robot’s body. Elevation samples provides a convenient exteroceptive input of the policy network.

\paragraph*{Action Space}
The policy network in our system outputs a 16-D gait schedule parameters consisting of target phase and joint residual angles for four legs. 
The gait schedule parameters are then fed as an input to the policy modulated trajectory generator (PMTG) module \cite{ICTZCSV18}, which generates the 12-D dof target positions. 
The inclusion of PMTG enhances the training of its upstream policy network more stable and sample-efficient, resulting in regular gait movement.


\subsection{Reward shaping}
\begin{figure}[htbp]
  \centering  
  \includegraphics[width=0.47\textwidth]{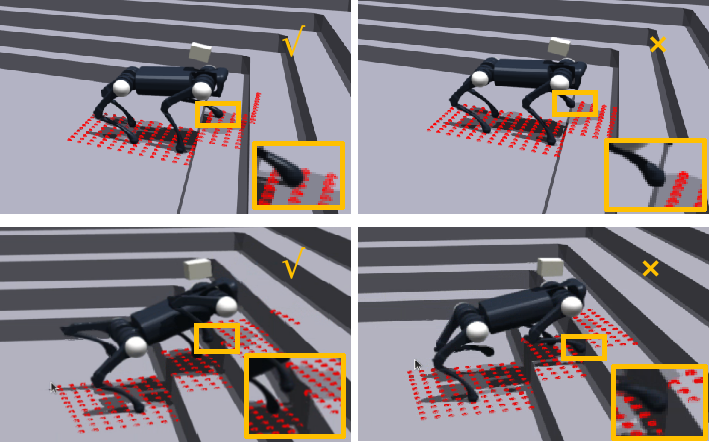}
  \caption{Behaviour-tuning rewards: the top two subplots illustrate the effects of the ``foot stance'' reward, which encourages the quadruped to place its footholds away from the edges of stair treads; the bottom two subplots illustrate the ``stumble'' reward effect, which prevents stumbled legs.}  
  \label{fig:reward}  
\end{figure}

The reward functions used for policy training are extended from the ones in \cite{RHRH22}; see Table \ref{table:reward} for a comprehensive list. The same set of rewards is applied to all sub-terrains in the experiments. Among all reward terms, the most prominent ones are linear and angular velocity tracking, which yields an instruction-following policy. The rest of the reward terms are engineered to finetune the walking behavior such as posture and foothold planning, as shown in Fig. \ref{fig:reward}. 

\begin{table*}[t]  
\centering  
\caption{Reward shaping.}  
\begin{tabular}{lll}
    \hline
    Reward & Formulation & Weight \\
    \hline
    linear velocity tracking  & $\exp(-\|^{B}\bm{v}_{xy}-^{B}\bm{v}_{xy}^{target}\|^2/0.25)$   & 2.0  \\
    angular velocity tracking & $\exp(-|^{B}\bm{\omega}_z-^{B}\bm{\omega}_z^{target}|^2/0.25)$ & 1.0 \\
    velocity constraint  & $-|^{B}\bm{v}_z|^2 -\|^B\bm{\omega}_{xy}\|^2/4$  & 2.0  \\
    orientation               & $ -0.1 \cdot \Theta_{pitch}^2 - \Theta_{roll}^2 $  & 5.0   \\
    base height               & $-\sum_{i=1}^4 {c_{i}\cdot [h^{des} - |h_{s_i} - h_{f_i}|]_+}$      & 1.0  \\
    torque                    & $-\sum_{i=1}^{12}|\bm{\tau}_i|^2 $            & 0.0002    \\
    dof acceleration          & $-\sum_{i=1}^{12} (^{t}\dot{\bm{\theta}_{i}} - ^{t-1}\dot{\bm{\theta}_{i}})/ \delta t $  & 1.25e-7 \\
    target smoothness         & 
    \tabincell{l}{
        $-\sum_{i=1}^{12} [({^{t-1}\bm{\theta}_{i}^{target} - ^{t}\bm{\theta}_{i}^{target}})^2 $ \\
        \qquad $+ ({^{t}\bm{\theta}_{i}^{target} - 2 \cdot ^{t-1}\bm{\theta}_{i}^{target}} + ^{t-2}\bm{\theta}_{i}^{target})^2]$
    }
    &   0.01 \\
    stand still & $ -\sum_{i=1}^{4}{|\theta_{i}^{abad}|}  $ & 0.5 \\
    dof pos limits            & $ \sum_{i=1}^{12}(-[\bm{\theta}_{i} - \theta_{max}]_+ - [\theta_{min} - \bm{\theta}_{i}]_+) $  & 10.0 \\
    collision                 & $-\sum_{body\in\{base, shoulder, legs\}}\mathds{1}[body \enspace in \enspace contact]$         & 1.0 \\
    foot air time             & $ -\sum_{i=1}^4{\mathds{1}[foot_i \enspace first \enspace contact]\cdot(T^{sw}_{i} - T^{des})}$ & 1.0 \\
    foot stumble & $ -\sum_{i=1}^4{\bm{f}_{i,xy} > 5 \cdot \bm{f}_{i,z}}$ & 0.5 \\ 
    foot stance               & $-\sum_{i=1}^4 c_i \cdot \sum_{j \in \mathcal{N}(i)} w_{i,j}  |h_{j} - h_{i}|)$
     &  5.0 \\
    foot height               &$ -\sum_{i=1}^4 {\mathds{1}[foot_i \enspace swinging]\cdot [h_{i} - h_{i}^{des}]_+}$                         & 3.0 \\
    proximity constraint     & $ -\| \delta \bm{\phi} \| -\| \delta \bm{\theta} \|$  & 0.1 \\
    \hline
\end{tabular}  
\label{table:reward} 
\end{table*}

\subsection{Zero-shot Transferable Simulation}

\paragraph*{Domain randomization}
Domain randomization as a key ingredient for zero-shot sim-to-real transfer remedies the sim-to-real gaps. We intend for a robust control policy by randomizing the following components:
\begin{enumerate}
    \item Intrinsic physical parameters: mass offset and center-of-mass (CoM) shift of the robot and its onboard sensor, if any.
    \item Extrinsic physical parameters: contact friction coefficient, restitution coefficient.
    \item Actuator parameters: motor strength, PD control coefficients.
    \item Proprioceptive observations: noises on base linear and angular velocities, dof position and velocity.
    \item Exteroceptive observations: nominal noise is added to all height samples in the vertical direction; drift noise is added to the root position affecting all height samples at once, which emulates potential localization error.
\end{enumerate}

\noindent
Table \ref{table:randomization} provides the complete list of randomized parameters and their variation ranges. Actuator parameters, intrinsic and extrinsic physical parameters, are randomly drawn once at the beginning of the simulation and stay fixed during training. Proprioceptive and exteroceptive observation noises are generated every time step.

\begin{table}[htbp]
    \centering
    \caption{Domain randomization}
    \begin{tabular}{cc}
    \hline
    Items & Range \\
    \hline
    base mass offset (kg) & [-2, 2] \\
    base CoM shift (m) & [-0.05, 0.05] \\
    Lidar mass offset (kg) & [-0.3, 0.3] \\
    Lidar CoM shift (m) & [-0.02, 0.02] \\
    friction coefficient & [0.5, 1.25] \\
    restitution coefficient & [0.4, 0.6] \\
    motor strength factor$^*$ & [0.8, 1.2] \\
    $K_p$ factor$^*$ & [0.75, 1.25] \\
    $K_d$ factor$^*$ & [0.5, 1.5] \\
    base linear velocity noise (m/s) & [-0.1, 0.1] \\
    base angular velocity noise (rad/s) & [-0.2, 0.2] \\
    dof position noise (m) & [-0.01, 0.01] \\
    dof velocity noise (m/s) & [-1.5, 1.5] \\
    height measurement noise (m) & [-0.02, 0.02] \\
    root position noise$^*$ (m) & [-0.05, 0.05] \\
    \hline
    \end{tabular}\\[4pt]
    \label{table:randomization}
\end{table}

\paragraph*{Perception Alignment}
We bridge the perception gap between simulation and real world by employing pre-processing techniques and 3D reconstruction, as shown in Fig. \ref{fig:perception}. Specifically, for the depth camera fixed to the front of the robot’s head, capturing areas including the front legs, we simulate leg motion noise, Gaussian noise, and random artifacts in elevation samples during simulation. Furthermore, we translate the point cloud to the body frame and perform depth hole-filling in the real world. Additionally, for the solid-state Lidar, we provide Lidar-IMU fused odometry and construct the terrain mesh to sample the elevation beneath the body frame.

\paragraph*{Terrain curriculum} 
Within the simulation environment, we construct a training playground consisting of $10 \times 10$ grid cells, where each of the ten rows corresponds to a subterrain type and splits into ten levels with increasing difficulty.
When training control policies on this playground, a game-inspired curriculum learning is enacted to smooth the learning process \cite{MLHWKH22,RHRH22}. That is, an agent is promoted to a neighboring higher level of the same row if it traverses sufficiently long distance before being reset.
\section{IMPLEMENTATION}
\subsection{Deployment on quadruped robot}
Our approach achieves zero-shot transfer from simulation to reality on the Unitree A1 and Aliengo quadruped robots. We mount a solid-state Lidar, LIVOX Mid-70, on the head of A1, and a compact and lightweight depth sensor, Realsense D405, on Aliengo, for terrain perception and reconstruction. The elevation samples are acquired from the depth camera at 30 Hz or from Lidar construction at 10 Hz. Our control policy runs forward inference at 80 Hz, and we send joint angle commands while updating the degrees of freedom (DoF) states at 320 Hz. The PMTG defaults to the trotting gait at a base frequency of 1.5Hz. We utilize an RTX 3080Ti (with 16GB RAM) and ISAAC robot simulation platform for reinforcement learning training, and deploy the trained policy network on Ascend development board, which is accelerated by Compute Architecture for Neural Networks (CANN). Our control policy is trained exclusively in simulation and deployed to the quadruped robot in reality without further fine-tuning efforts.


\subsection{Challenges}
\paragraph*{Locomotion under challenging conditions}

\begin{figure*}[htbp]
  \centering  
  \begin{tabular}{ccc}
  \includegraphics[width=0.305\textwidth]{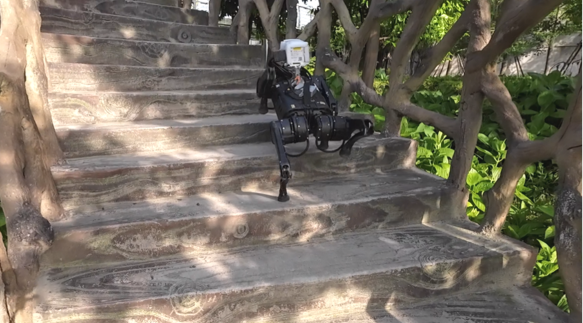} & 
  \includegraphics[width=0.305\textwidth]{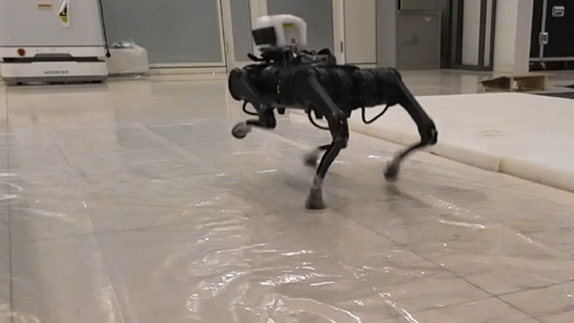}
  & 
  \includegraphics[width=0.305\textwidth]{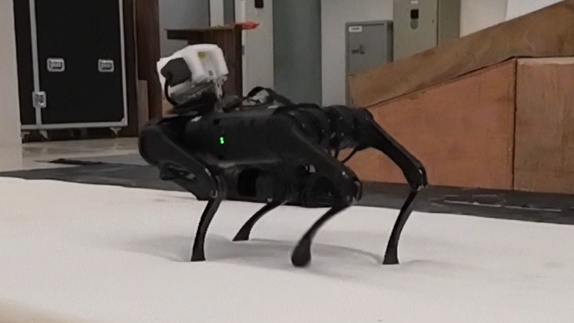}\\
  Stairs & Oily Surface & Deformable \\[2pt]
    \includegraphics[width=0.31\textwidth]{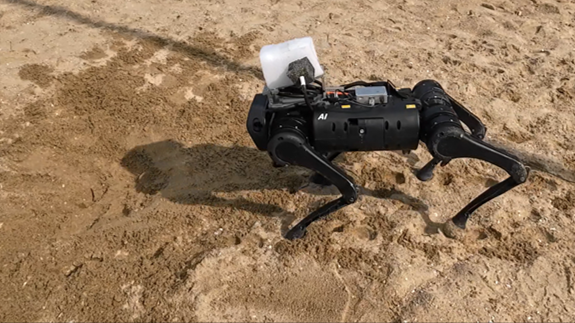} & 
  \includegraphics[width=0.305\textwidth]{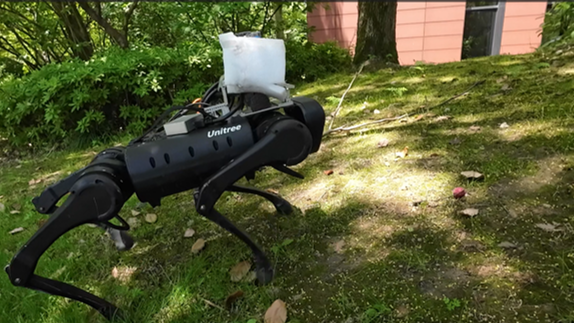}
  & 
  \includegraphics[width=0.305\textwidth]{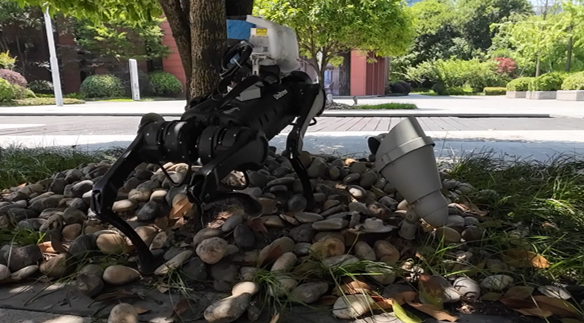}\\
  Sand & Slope Lawn & Cobblestone
  \end{tabular} 
  \caption{Selected terrains for real-world evaluation.}  
  \label{fig:benchmark1}  
\end{figure*} 

    
We have selected a wide range of challenging indoor and outdoor environments for experimental validation, which include stairs, oily surfaces, deformable ground, sandy fields, rough lawns and cobble stone. These terrains vary wildly in their material properties which are never experienced during training in simulation, and often carry random elevation changes easily misinterpreted by Lidar or depth sensor. 
With our sensor setup, exteroceptive challenges such as restrictive field of view and reflection noise are foreseeable.

\paragraph*{Hardware Inconsistency}
Besides, the actuator hardware and IMU can possibly deteriorate over time or start off with miscalibrations, to a less or more extent depending case by case. In a more severe case, one can observe that the (same) built-in controller yields on visibly different movement on different robots. While such hardware caveats can be fixed by dedicated technicians, we stick to our promise of zero-shot transfer. That being said, we accept possible hardware caveats as they are and strive for a remedy by learning a sufficiently robust control policy.

Our transferred control policy has consistently shown strong empirical performances across all tested scenes. 
Meanwhile, state-of-the-art imitation-based approaches, namely teacher-student privileged learning \cite{MLHWKH22} and rapid motor adaptation \cite{kumar2021rma} experienced difficulties especially under strong domain randomization. Policies from imitation or adaptation learning exhibit visible degradations such as non-zero velocity offset, body tilt, or stumbling through stairs.



\subsection{Experiments}
\paragraph*{Real-world quantitative evaluation}

We elaborate our experimental validation with quantitative evaluation in the real world, as shown in Fig. \ref{fig:benchmark1}. These test scenes include stairs with 15cm-high steps, plastic film slippery ground poured with olive oil, sandy ground, loose sponge mats with varying degrees of variability, outdoor lawns, and piles of pebbles. In a more controllable setup, we maneuver the robot deployed with each candidate policy to traverse multiple scenes over a distance of 5--10 meters repeatedly and record its success rate per scene after five trial runs. The results are shown in Table \ref{tabel:real_benchmark}. T-S = Teacher-Student \cite{MLHWKH22}; RMA = Rapid Motion Adaptation\cite{kumar2021rma}; MPC = Model Predictive Control (pre-installed in Unitree robots). Weak randomization refers to switching off randomization on motor strength, $K_p$, $K_d$, and root position noise marked in Table \ref{table:randomization}.

\begin{table}  
\centering
\caption{Success rate of Real-world evaluation.}  
\begin{tabular}{cccccc}
    \hline
     Method & Ours & Ours         & T-S           & RMA  & MPC \\
    &           & (weak rand.)  &     \\
    \hline
    \multicolumn{6}{c}{Success Rate (\%)}\\
    \hline
    Stairs & \textbf{100} & 40 & 60   & 0 & 80\\ 
    Oily Surface & \textbf{80} &  40 & 20  & 40 & 0  \\
    Deformable & \textbf{100} & 60 & 60  & 40 & 20 \\
    Sand & \textbf{80} & 60 & 60  & 20 & 20\\
    Slope Lawn & \textbf{100} & 80 & 80  & 60 & 80\\
    Cobble stone & 100 & 100 & 100  & 100 & 100\\
    \hline
\end{tabular} \\

\label{tabel:real_benchmark}
\end{table}  

Our method is consistently more successful across all test scenes. As important empirical evidence, we switch off selected randomizations related to actuator and localization (termed weak randomization) and found that the extended domain randomization provides a vital boost in real-world performance, with end-to-end RPPO training. The teacher-student style training, namely ``T-S'' and ``RMA'', suffers from imitation gap under stronger domain randomization. We also compared to a model predictive control baseline (without RL) whose performance fluctuates from one subterrain to another.


\paragraph*{Metric evaluation in simulation}
\begin{table}  
\centering 
\caption{Quantitative evaluation in simulation.}
\begin{tabular}{ccccc}
\hline  
    Method & Ter. Level & Distance & Torque & Suc. Rate (\%)\\
    \hline
    Ours & \textbf{8.9} & \textbf{72.7} & 37.2 & \textbf{89}\\
    Ours (weak rand.) & 7.9 & 64.6 & 37.2 & 79 \\
    T-S  & 7.4& 60.9 & \textbf{36.0} & 74 \\
    RMA  & 7.9 & 65.1 & 38.2 & 77.5\\
    \hline
\end{tabular} \\[4pt]
\label{label:2}
\end{table}  
We complement our studies in simulation by evaluating metrics that are difficult to measure in reality and at larger scales.

Evaluation is conducted on playing trained policies on a $10$-by-$10$ grid-style playground, similar to the training environment but built with more challenging parameters.
For instance, the step heights for the stairs in evaluation ranged from 3cm to 18cm, compared to the range between 2cm and 12cm during training.
All domain randomization and noises listed in Table \ref{table:randomization} are activated. 
We evaluated according to four metrics in Table \ref{label:2}, when 200 parallel robots traversed their respective subterrains from lowest to highest levels in one attempt. 

The results in Table \ref{label:2} again provide consistent evidence that our method is the most performant among others, although the gaps between different methods appear more narrow in the simulated world. With our method trained under weak domain randomization (yet tested under strong domain randomization), one can already observe performance gaps with manually prescribed distributional drifts --- the gaps are only expected to widen during sim-to-real transfer. Overall, the quantitative evaluation in simulation gives a reasonable projection of how different policies shall perform when being transferred to the real world.

\paragraph*{Domain transfer consistency on hardware}
\begin{figure}[htbp]
  \centering   
  \includegraphics[width=0.47\textwidth]{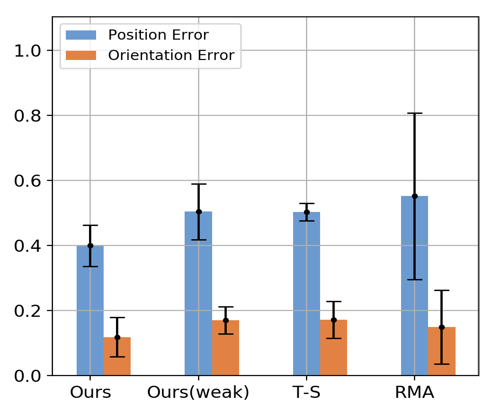}\\  
  \caption{Behavior consistency comparison: statistics of position and orientation errors.}  
  \label{fig:behavior}  
\end{figure}  
Upon completion of training in simulation, control policies are transferred to the real world for deployment. Discrepancies in motion phenomena arise when deploying the same policy across different robot hardware. In this experiment, we compared our control policy with teacher-student control policies on three A1 robots. Specifically, we measured means and standard deviations of the position drift errors (Euclidean distance between the initial and final positions) and orientation errors (Euclidean distance between the initial and final Euler angles) of each robot in five runs. The robots were commanded to perform a stationary stepping motion on the ground for a duration of 10 seconds, while a motion capture system is employed to detect any position or orientation drift.

The summarized data is visualized in Fig. \ref{fig:behavior}. The results indicate that our control policy achieved the lowest position and orientation errors among the tested policies, thereby demonstrating minimal behavioral discrepancies between the virtual and real-world domains. It is important to note that variations in motor dynamics and IMU miscalibration inevitably lead to diverging behaviors among the A1 robots. Nevertheless, our policy exhibited the highest level of consistency across different A1 robots compared to the other evaluated policies.

\begin{figure}[t]
  \centering   
  \includegraphics[width=0.47\textwidth]{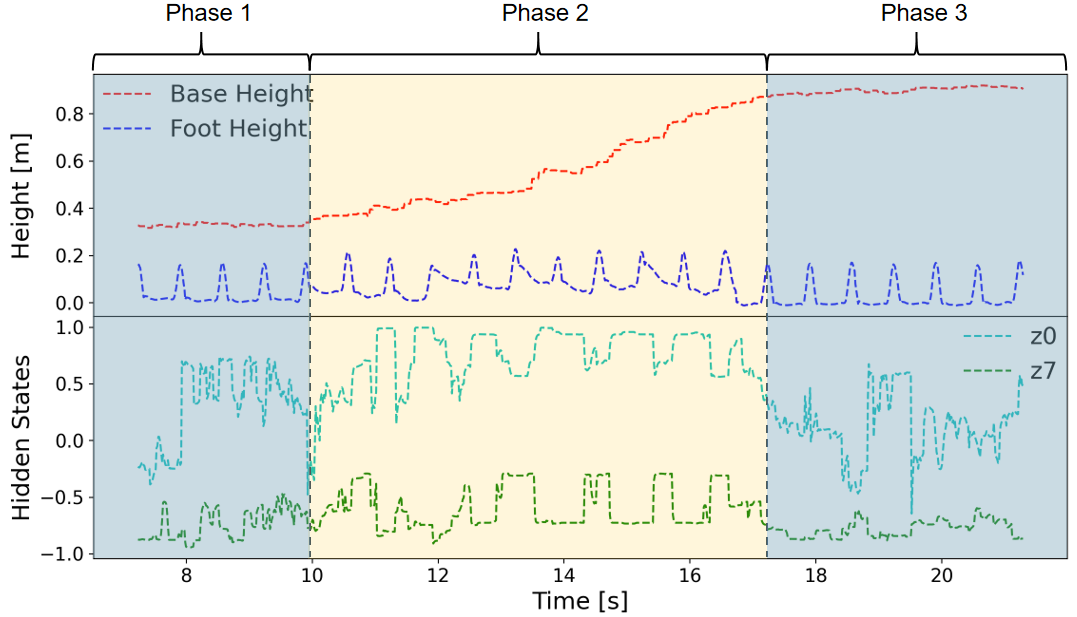}
  \caption{Introspective analysis on climbing stairs. The whole process consists of three phases: (1) approach the stairs; (2) move up the stairs; (3) recover on the flat platform.} 
  \label{fig:analysis}  
\end{figure}

\paragraph*{Introspective analysis} We conducted an introspective analysis of our control policy for a simple task of climbing stairs, as depicted in Fig. \ref{fig:analysis} (top). The whole process consists of three phases: (1) approach the stairs; (2) move up the stairs; (3) recover on the flat platform. The base height illustrated in the upper plot is approximately 28cm in Phase 1. The blue curve presents the z-coordinate of the front-right foothold in a world-aligned base frame offset by 28cm. Base and foot heights in the plot are measured in meters. 
Our policy generates actions that result in higher foothold lifting compared to Phase 1 and Phase 3, suggesting that the learned policy indeed plans the foot trajectories based on terrain and body orientation observations.

\begin{figure}[t]
  \centering   
  \includegraphics[width=0.47\textwidth]{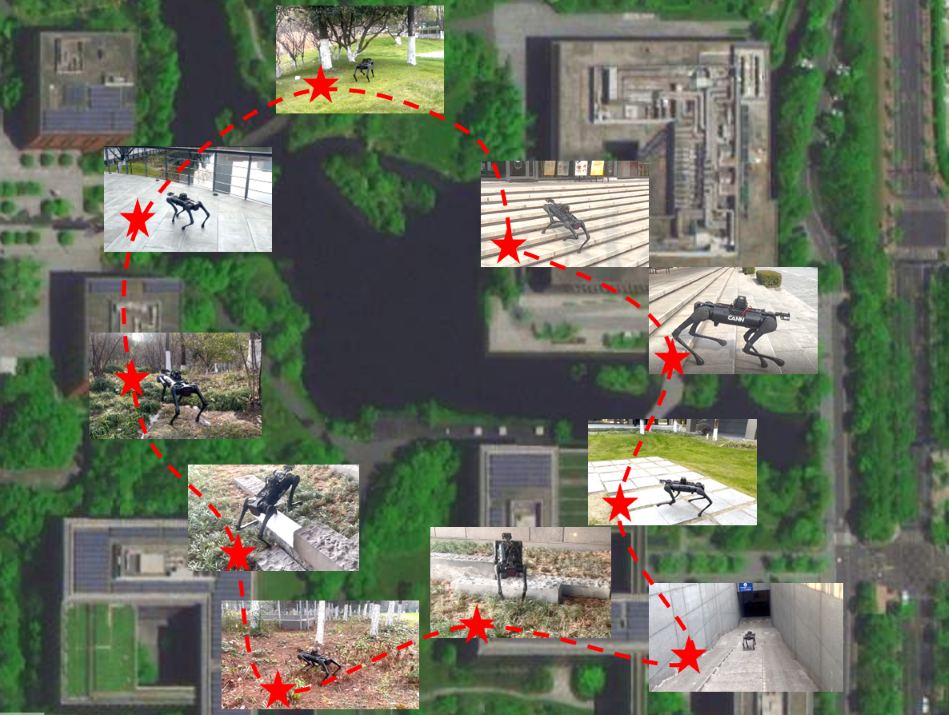}
  \caption{Campus inspection with Aliengo which mounts a depth camera on the head.}  
  \label{fig:inspection}  
\end{figure}  
To gain insights into the inner mechanism of our recurrent policy, we plot the internal state profiles within the GRPN network in Fig. \ref{fig:analysis} (bottom). Notably, the first and eighth components $z_0, z_7$ of the state in the final RNN layer exhibit pattern variations during phase transitions. The pattern observed during the second stage differs from the other two phases. With further experiments, we found that policies that did not incorporate terrain observations failed to traverse the stairs. This evidence empirically supports that our policy network is indeed influenced by terrain observations.

\paragraph*{Application of campus inspection} 
We showcase a real-life application of campus inspection with Aliengo which mounts a depth camera on the head to sample data from the area beneath the head and front legs. The task covers a total distance of approximately 2 kilometers and takes about an hour. The quadruped robot succeeds in walking through various types of terrain without manual assistance, including stairs, slopes, stone pathways, grassy field, and cobblestone place, as shown in Fig. \ref{fig:inspection}.

\section{CONCLUSIONS}

We have presented a robust and general end-to-end approach for quadrupedal locomotion using zero-shot learning architecture incorporating a new recurrent proximal policy optimization algorithm (RPPO). Our approach is trained in simulation and directly deployed on the real robots without further fine-tuning. This significant reduces maintenance costs, and enhances deployment efficiency and scalability. We have performed extensive experiments and comparison in many challenging terrains and the results showed significant performance improvement against existing methods. In future, we are going to extend our approach to handle discontinuous terrains like gullies. 

\section*{ACKNOWLEDGEMENTS}
We would like to express our deep gratitude to Yizhen Chen who have contributed to the completion of this paper.

\bibliographystyle{IEEEtran}
\bibliography{ref}

\end{document}